\documentclass[sigconf]{acmart}
\AtBeginDocument{%
  \providecommand\BibTeX{{%
    \normalfont B\kern-0.5em{\scshape i\kern-0.25em b}\kern-0.8em\TeX}}}

\acmSubmissionID{549}

\begin{document}

\title{Sketch Me A Video}




\author{Haichao Zhang}
\affiliation{%
  \institution{Zhejiang University}
  \city{}
  \country{}
  }
\email{zhanghaichao@zju.edu.cn}

\author{Gang Yu}
\affiliation{%
  \institution{Tencent}
    \city{}
  \country{}
  }
\email{skicyyu@tencent.com}

\author{Tao Chen}
\affiliation{%
  \institution{Fudan University}
    \city{}
  \country{}
  }
\email{eetchen@fudan.edu.cn}

\author{Guozhong Luo}
\affiliation{%
  \institution{Tencent}
    \city{}
  \country{}
  }
\email{alexantaluo@tencent.com}

\renewcommand{\shortauthors}{Haichao Zhang, et al.}

\begin{abstract}
   	Video creation has been an attractive yet challenging task for artists to explore. With the advancement of deep learning, recent works try to utilize deep convolutional neural networks to synthesize a video with the aid of a guiding video, and have achieved promising results. However, the acquisition of guiding videos, or other forms of guiding temporal information is costly expensive and difficult in reality. Therefore, in this work we introduce a new video synthesis task by employing two rough bad-drwan sketches only as input to create a realistic portrait video. A two-stage Sketch-to-Video model is proposed, which consists of two key novelties: 1) a feature retrieve and projection (FRP) module, which parititions the input sketch into different parts and utilizes these parts for synthesizing a realistic start or end frame and meanwhile generating rich semantic features, is designed to alleviate the sketch out-of-domain problem due to arbitrarily drawn free-form sketch styles by different users. 2) A motion projection followed by feature blending module, which projects a video (used only in training phase) into a motion space modeled by normal distribution and blends the motion variables with semantic features extracted above, is proposed to alleviate the guiding temporal information missing problem in the test phase. Experiments conducted on a combination of CelebAMask-HQ and VoxCeleb2 dataset well validate that, our method can acheive both good quantitative and qualitative results in synthesizing high-quality videos from two rough bad-drawn sketches.   
    
\end{abstract}



\maketitle

\section{Introduction}

\begin{figure}[t]
\begin{center}
\includegraphics[width=0.9\linewidth]{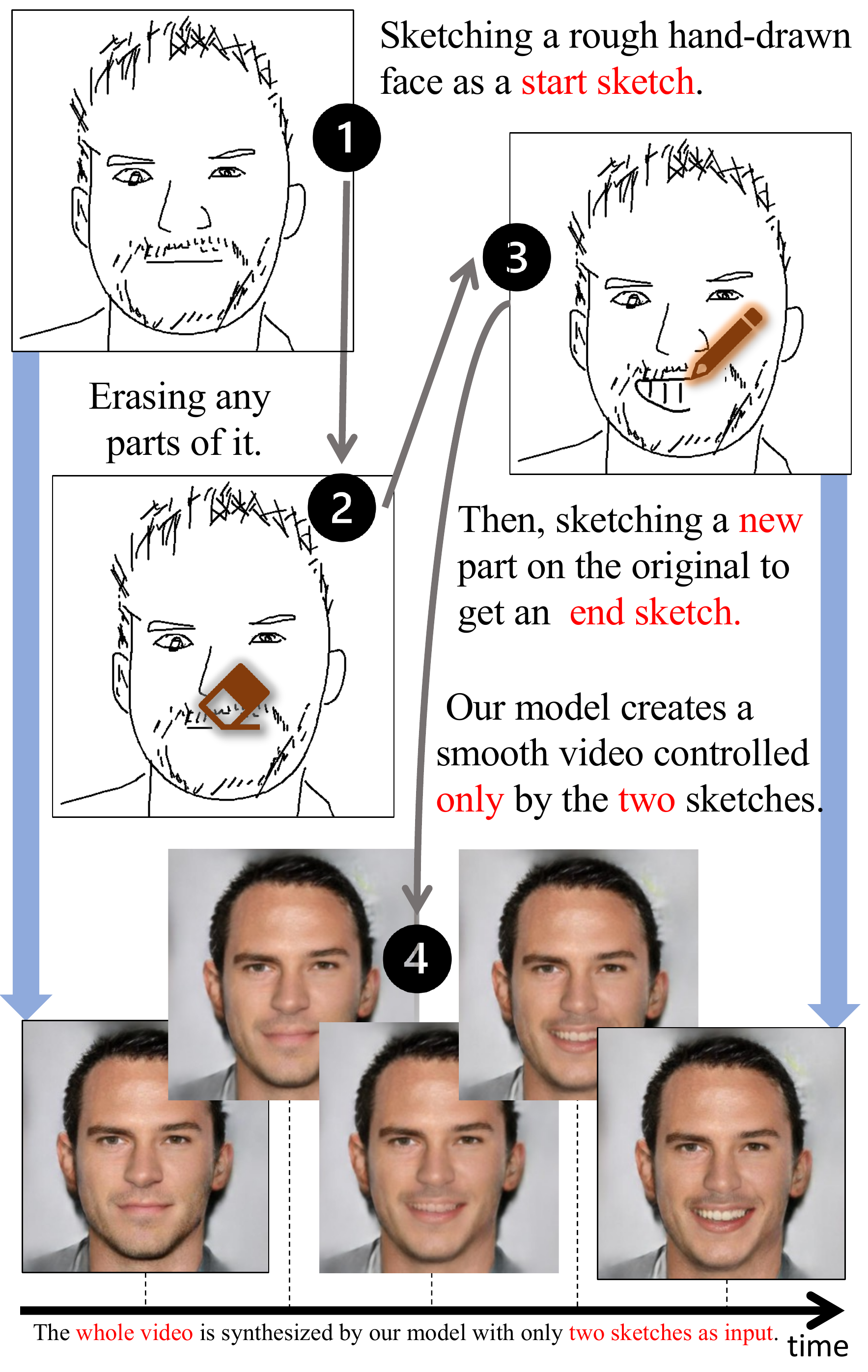}
\end{center}
   \caption{One illustrative example on how our sketch-to-video task is used in practice. Our model requires only two rough hand-drawn sketches as input to synthesize a natural and smooth video. It could work interactively and iteratively, as shown in the figure. }
\label{fig:intro}
\vspace{-1em}
\end{figure}
Video creation has often been considered as a professional task in the animation industry, which can only be done by well-trained experts. Recent progress on artificial intelligence makes it easy for researchers to generate realistic videos via deep learning models, e.g., the generative adversarial networks (GANs). However, it is hard for general users and artists to create videos by directly utilizing these deep models. As shown in Fig.1, in this work, we aim to do a new task which creates a video from two input sketches that are generated by simply modifying a rough hand-drawn sketch image. The two rough hand-drawn sketches specify the start and end frames for the video to be synthesized.
	
We review existing literatures that are related to video creation task. We divide these works into three categories. The first category~\cite{Wang}~\cite{Wang2019}~\cite{Chen2019} is video translation, and video prediction, e.g., Vid2vid~\cite{Wang}, Fewshot-vid2vid~\cite{Wang2019}. which require a guiding label sequence as conditional input and impose high cost for such guide data collection. Our work is different from this line of works, by avoiding to use such high-cost label sequence as auxiliary input.

The second category is video interpolation~\cite{Shen2021,Yan2021,Park2021,Bao2021}, which translates a video with low FPS into a high FPS one. The interpolation operation generates more video frames from two consecutive frames, which assumes the motion variations between consecutive frames conform to a mathematical distribution, like linear variation(Super SloMo~\cite{Jiang2018}), quadratic variation (Quadratic Video Interpolation~\cite{Xu2019}), etc. Such assumption is suitable when two consecutive input frames are sampled from a video with very short interval (usually measured in milliseconds), ensuring that the two frames have very high similarity. However, the time interval between two input images is random and usually a few seconds in our task, which makes the motion variation to become very complex and causes the above mathematical distribution assumptions incorrect. 

The third catgory is video inbetweening, which generates plausible and diverse video sequences given only a start and an end frame, and is mostly closed to our work. Earlier works (Computer-aided inbetweening) often pay attention to computer-aided software design to generate simple shape frames, which is hard to handle realistic images and complex motions between rough hand-drawn sketches. The recent representative works are Point-to-Point Video Generation~\cite{Wang2019a} and Deep Sketch-guided Cartoon Video Inbetweening~\cite{Xiaoyu2020}. The former needs prior knowledge like pose skeletons, while the second one needs a well-drawn cartoon sketch and an actual movie frame as inputs. Even though these two methods achieved promising performance, but both the well-drawn sketches and prior knowledge such as pose skeletons used in these works are not easy to acquire in real use, especially in our scenario where only rough hand-drawn sketches are provided by common users. Therefore, our model is the first to synthesize a video with only two rough hand-drawn sketches by common users, which is a new task that faces new challenges not solved by existing works as follows.

The first challenge is translation from bad-drawn sketches to natural videos. As general users often draw the rough hand-sketch in a casual way, which may result in no image edges. Furthermore, the sketching styles vary a lot for different people, causing the drawn sketches to present a highly free-form distribution. It is extremely difficult to build a dataset to cover all these distributions, thus making the bad-drawn sketch-to-video translation an out-of-domain problem.

The second challenge is for our task is without guiding temporal information. It is impossible to ask general users to draw a sketch for each video frame to be synthesized. Without temporal information, the motion of generated video could come from nowhere and beyond control. Current works ~\cite{Wang}~\cite{Wang2019}~\cite{Chen2019} focus on the video translation with conditional input sequences, which contain the guiding temporal information. Some works~\cite{Wang2019a} try to synthesize the conditional sequences by leveraging prior knowledge, and others~\cite{Xiaoyu2020} ~\cite{Xu2019} even directly calculate the motion based on mathematical assumptions. Due to the arbitrary and flexible input given by users, neither prior knowledge nor mathematical assumptions can we use to synthesize guiding temporal information in our scenario.

To this end, we propose a new pipeline to address the above challenges. We find that it is difficult to design an end-to-end model which can synthesize videos undistortedly and meanwhile translate the bad-drawn sketches to natural images, due to the out-of-domain problem mentioned above. Therefore, we design our sketch-to-video model from two stages: (1)Stage-I addresses the out-of-domain challenge via the developed feature retrieval and projection (FRP) module. The FRP module maps the sketch facial parts into semantic space and retrieves serval closest ones in the training dataset, then projects them to get a reasonable semantic feature sequences, sent to stage-II to create videos. (2)Stage-II addresses the temporal information problem from several aspects. Firstly, a temporal motion projection (TMP) module is designed to project the training videos into a motion space conforming to a standard normal distribution. Note that the TMP module is not used \textbf{in testing} and the motion variable is sampled from the standard normal distribution. Secondly, a feature blending (FB) module is developed which combines the motion variable with the generated semantic features in Stage-I to reconstruct the final video. In addition, as a smooth motion from the initial sketch to the end one is required for video synthesis but the motions in the video dataset present various unpredictable directions, we thus design a smooth motion loss to penalize the unwanted direction while keeping the wanted motions to be natural and smooth. 

In this work, we take the front face sketches as examples to introduce how to synthesize face videos from rough sketches. The contributions are summarized as follows:
\begin{itemize}
    \item A new sketch-to-video task, that helps general users create videos easily with only two rough hand-drawn sketches, is proposed for the first time.
    \item A feature retrieval and projection (FRP) module which can alleviate the out-of-domain problem due to arbitrarily drawn sketches, is designed to help to translate a bad-drawn sketch to a natural video.
    \item A motion projection followed by feature blending method is proposed, which can alleviate the guiding temporal information missing problem in our task.
\end{itemize}

\section{Related Works}
\subsection{Video Creation.}
We have reviewed existing works related to video creation task, and find that this is the first work to do sketch-to-video generation from only two given rough bad-drwan sketches. Nevertheless, we still present three kinds of video generation works which are mostly related to ours, as described before, which are divided into video translation ~\cite{Wang}~\cite{Wang2019}~\cite{Chen2019} and prediction \cite{Wang}~\cite{Wang2019}, video interpolation~\cite{Shen2021,Yan2021,Park2021,Bao2021},  and video inbetweening ~\cite{Wang2019a}~\cite{Xiaoyu2020}. These works have been analyzed in the previous section and are all different from our work which rely on rough sketches only. We next will give some review on sketch processing. 
\subsection{Free Sketch Processing.}
Free-hand sketches are highly illustrative, and have been widely used by humans to depict objects or stories for a long time. The recent popularization of touchscreen devices has made sketch creation a much easier task than ever and
consequently made sketch-oriented applications increasingly popular. The progress of deep learning has immensely benefited
free-hand sketch research and applications. However the sketches could be divided into two categories. The first one takes well-drawn sketch which is usually got via edge detection from drawings or from artists, which has stable distribution and could be easily translated. The second category is free sketch drawing (Skecthy GAN~\cite{Chen2018}, Sketh me that shoe~\cite{Yu2016}), whose distribution is too causal and free-form, and lacks structure mapping between the sketch and object in real world. That makes it hard to expolre. Benefiting from the
development of feature engineering, the early works use image retreive method to find a closest sample in semantic space. Direcly translating those sketches into multimedia domain, could cause distortion. Therefore, temporal information is often introduced to help synthesize the video.

\subsection{Temporal Video Guide}
Most video creation works ~\cite{Wang}~\cite{Wang2019}~\cite{Chen2019} do video translation based on conditional input sequences, which contain the guiding temporal information serving as auxiliary input to help synthesize the target video.
For example, video interpolation~\cite{Wang2018}~\cite{Jiang2018} as a classical vision problem aiming to synthesize high FPS videos from a low one, can synthesize serval more frames from given two based on given temporal reference. The SuperSlomo~\cite{Jiang2018} as a famous video interpolation method which has been widely used in video creation, also leverages on extra guide input to interpolate more frames.  

When no conditional input is available, some works~\cite{Wang2019a} try to synthesize the conditional sequences as auxiliary input by employing prior knowledge such as object normal behaviours, daily activities, etc. There also exist some works~\cite{Xiaoyu2020} ~\cite{Xu2019} which even directly calculate the motion changes based on certain mathematical distribution assumptions, and leverage on the derived motions to help synthesize the video. Our work on the contrary does not utilize any temporal sequence or any other form of guide information, and purely use the rough sketches to create the video. 

\section{Method}
\begin{figure*}[t]
  \includegraphics[width=\linewidth]{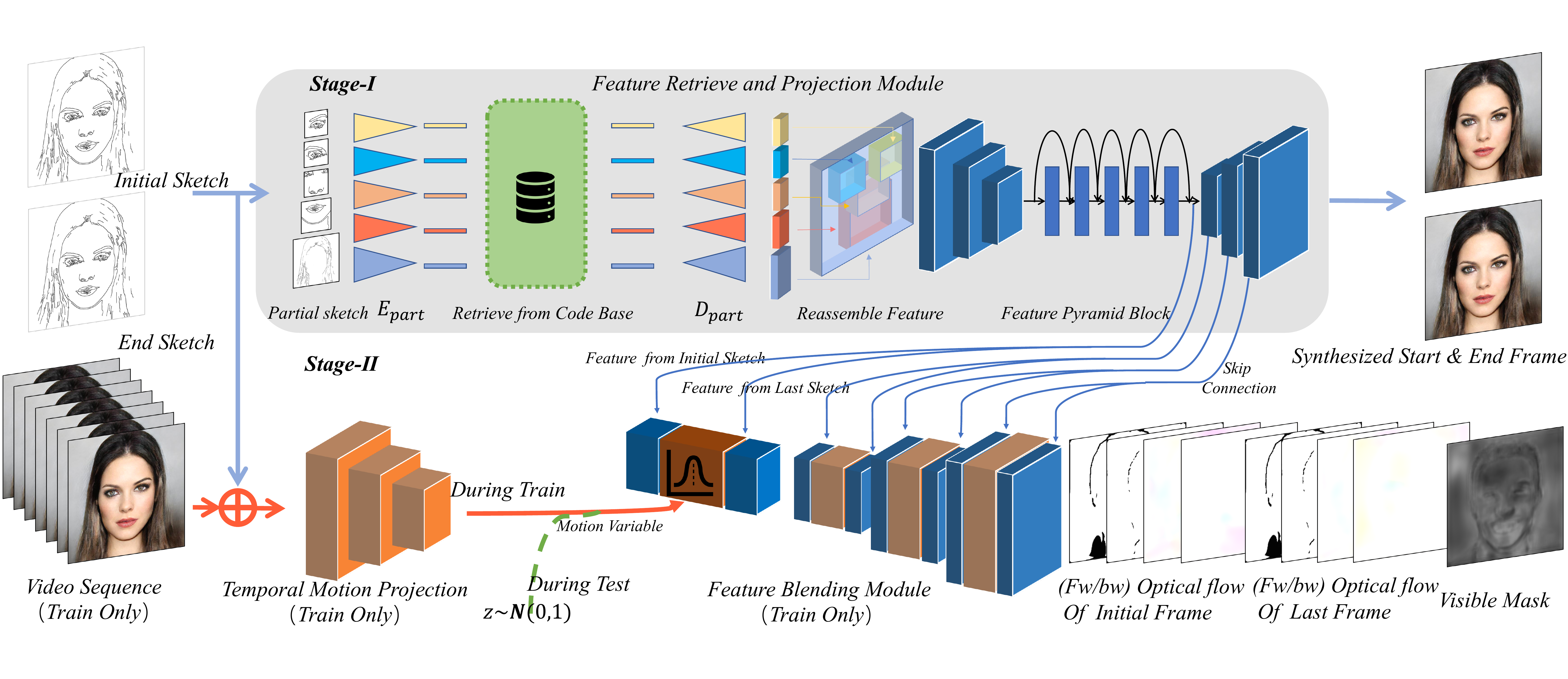}
  \caption{The overall architecture of the proposed sketch-to-video generation network from bad-drawn sketches. It consists of two stages: a) Feature Retrieve and Projection, and b) Motion Projection and Video Generation. Given two rough hand-drawn face sketches, the extracted semantic face features by Stage-I and projected motions by Stage-II are embedded and blended together to synthesize a natural and gradually changing video.}
  \Description{Enjoying the baseball game from the third-base
  seats. Ichiro Suzuki preparing to bat.}
  \label{fig:arch}
\end{figure*}
\subsection{Overview}
Our goal is to generate a smooth video from the given two bad-drawn face sketches in an interactive manner. The video should have the natural motion and meanwhile the attributes of the sketch such as identity, texture, etc. 

\textbf{Our Pipeline}.
We divide this challenging task into two stages:
\begin{itemize}
    \item \textbf{Stage-I model}: It takes the two bad-drawn sketches to generate two reasonable start \& end frames, and meanwhile produce the intermediate semantic feature sequences to be used in Stage-II.
    \item \textbf{Stage-II model}: It takes semantic feature sequences from the Stage-I model and blends them hierarchically with a motion variable extracted from the TMP module (in training phase) or the standard normal distribution (in the testing phase) to synthesize the final video.
\end{itemize}

\textbf{Symbols Definition.}
We denote the training video and the output video as $v$ and $\hat{v}$, respectively. The frame number of the two videos is $N$, and each frame in $\hat{v}$ is denoted by $\hat{v}_o, o\in [1, N]$. The start sketch and end sketch are denoted by $s_s$ and $s_e$, respectively.

\subsection{Feature Retrieve and Projection Module}
\label{FRP}
In Stage-I, we train an FRP module that takes input facial sketches to generate realistic frames and hierarchical semantic feature sequences, as shown in Fig.\ref{fig:arch}. 

Because the input of our Stage-I model is bad-drawn sketches, the developed FRP module is expected to tackle the out-of-domain challenge mentioned before. We train several partial encoders denoted by $E_{part}$ to project the sketch into semantic manifolds. The $E_{part}$ embeds different sketch parts into different embeddings seperately, which renders the model to be able to disentangle various facial attributes such as mouth, eye, nose, for subsequent feature retrieve and projection. 

To retrieve for the input sketch part feature, we need to build a codebase consisting of rich semantic embeddings of various facial skech parts by using the $E_{part}$ to encode the whole dataset. Then, for each input partial sketch, we can use the K-nearest neighbor algorithm to retrieve the top K closest embeddings which may represent the out-of-domain sketches in the semantic space of facial parts sketches. We follow the locally linear embedding (LLE)~\cite{Roweis2000} algorithm to project the Top-K samples into the embeddings of the out-of-domain sketch. 

After getting the partial sketch embeddings, we still need to design a module to get high-quality images (e.g., start and end) and meanwhile map those partial sketch embeddings into the semantic space of natural images. Hence, we adopt the widely used Pix2PixHD~\cite{Wang2018} as a backbone to do this task. However, we observe that directly decoding these partial embeddings into a realistic image faces the challenge that features of different facial parts are heavily entangled, which causes the difficulty of establishing a spcified atttribute manipulation mapping between input and output. To alleviate this, we exploit several partial decoders $D_{part}$ to reassemble the retrieved embeddings into an image-like feature pool, where each feature in the image corresponds to its original facial part (attribute) location in the input sketch. We then feed them into the Pix2PixHD~ backbone. Besides, we apply a feature pyramid block on the decoder of Pix2PixHD~\cite{Wang2018} backbone, which provides hierarchy information of different scales of the realistic face to benefit the video generation process in Stage-II.

We denote the whole process of Stage-I as:
\begin{equation}
    \hat{v_s},\hat{v_e},f_s,f_e = G_{frp}(s_s,s_e)
\end{equation}
where the $G_{frp}$ denotes the developed FRP module, and $f$ denotes a set of hierarchy features output by the feature pyramid block. Note that the number of features in a $f$ is equal to the number of the upsampling blocks. The subscript $s$ and $e$ denote the start and end frames, respectively. 

\subsection{Bad-drawn Sketch-to-Video Generation.}
In Stage-II, we propose a model that generates a video with only two bad-drawn sketches. The Temporal Motion Projection (TMP) module, used in training phase only, aims to project a training video sequence into a motion space that will be used in test phase. Then, the developed Feature Blending (FB) module fuses the motion variable from TMP and the semantic features from Stage-I in a disentangled way. The FB module generates a merging mask with bidirectional optical flows for both the start and end frames. Finally, the optical flow merging (OFM) module synthesizes the final video. Additionally, A smooth motion loss is designed to constrain the motion space to have only our desired directions by removing those unpredictable directions.

\textbf{Temporal Motion Projection Module.}
With the semantic features synthesized in Stage-I as the auxiliary, the challenge now is the lack of guiding temporal information between the sketches. We use the TMP module to project the training video into a standard normal distribution in training phase, while replacing the motion variable with a noise $z$ during the test phase to tackle this challenge. Leveraging the idea of cVAE~\cite{Xue2016}, we project the training video into a motion space. Specifically, we use multiple-layer perceptrons (MLP) to map the embeddings of TMP into mean and log variance to calculate the motion variable.  We use the Kullback-Leibler (KL) divergence as a loss function to make the distribution of the motion variables (referred to as \textbf{motion space}) conform to the standard distribution.

As the motion space from the training video may present weak relevance with the input sketches, the TMP module thus needs to consider encoding the motion between the input sketches to correlate the video motion with sketches. Besides, during the Stage-II training stage, the input sketches are directly from the start \& last frame edges of the training video v, and the training sketches are well generated through the edge detection method (HED~\cite{Xie2017}) with an edge simplification method(SimpleEdge~\cite{Simo-Serra2016}). We thus regard the sketch as a solid hint to the motion modelling, which benefits the optical flow generation. Therefore, we concatenate v with $s_s$, $s_e$ as the input of the TMP module. This process could be denoted as:
\begin{equation}
    f_m = E_{TMP}(z|v,s_s,s_e)~~~~~~~ , z \in N(0,1)
\end{equation}
Here we simplify the TMP module as an encoder $E_{TMP}$, and the $f_m$ is the motion variable. During the test, $f_m$ is sampled from standard distribution $N(0,1)$ and denoted as $z$.

\textbf{Feature Blending Module.}
After sampling a motion variable from the motion space generated by TMP, we need to further design afeature blending (FB) module to blend the motion variable $f_m$ with the extracted semantic features ($f_s$, $f_e$) from Stage-I via a video decoder. We concatenate the first item in $f_s$, $f_m$, and the first item in $f_e$ and feed the fused feature into the video decoder.
To complement the large scale information of targets, the upsampling block of the decoder also takes features from $f_s$ and $f_m$ at the same scale. 

The output of the FB module is optical flows with masks and a visible mask. The two pairs of optical flows for the start and end frames are bidirectional, where the reverse direction is to check the forward-backward consistency from the training video. The visible mask $m_v$ is a soft mask to coordinate the visibility of optical flows from the start and end frames.
This process could be denoted as:
\begin{equation}
    o_{ef},o_{eb},m_{ef},m_{eb},o_{sf},o_{sb},m_{sf},m_{sb},m_v = D_{FB}(f_s, f_e, f_e)
\end{equation}
where the $o$ is an optical flow sequence, and the $m$ is a mask sequence. The subscript $s$ and $e$ denote the flow is from the start and end frame, respectively. The subscript $f$ and $b$ represent the forward and backward flow, respectively.

\textbf{Optical Flow Merging Module.}
The optical flow merging (OFM) module generates the final video from the outputs of the FB module. To manage the flows from both the start and end frame and make the $\hat{v}$ stable, the OFM module merges the above optical flows and mask with following operation:
\begin{equation}
    \hat{v} = W(m_{eb}\odot o_{eb})\odot m_v+W(m_{sf},o_{sf})\odot(1-m_v)
\end{equation}
where the $W(\cdot,\cdot)$ is the warping process, and the $\odot$ denotes Hadamard product.
\subsection{Smooth Motion Loss.}
\label{sec: sml}
Although we can generate a natural motion space via the TMP~, the motion in the video dataset may be a random and free-form action beyond control. For example, in Fig.1, the input sketches are a man with a neutral mouth and a big smile mouth, respectively, but the corresponding motion in the training video could be started from a neutral mouth, then to blinking or talking for a whole, which does not conform to that of the bad-drawn sketches. In fact, the ideal motion is expected to present a gradual change mode from the start to the end sketch, in our bad-drawn Sketch-to-Video generation task. We have to find a method to control the generated motion to present similar distribution with the input sketches changes.  

An intuitive way is to directly introduce a constraint to the synthesized results, but this could cause the output to have a big gap with the ground truth, making the model hard to converge. To work on that, we carefully design a smooth motion loss to the optical flows $o_{ef}$ and $o_{sf}$ to make the motion grow gradually without abrupt changes.
The $\mathcal{L}_{sm}$ is formed as:
\begin{equation}
      \mathcal{L}_{sm} = o_{xf} - F_{li}(o_{xf}[0], o_{xf}[N]) ~~~~~~~, x\in{s,e}
\end{equation}
Here, the $o_{xf}[\cdot]$ is the index of the forward optical flow sequence $o_{xf}$, and the $F_{li}$ denotes the linear interlpolation. The ablation study~\ref{abla_res} in the experiment part well validates that after applying such a loss to the model, the unwanted motion disappeared and the motion change becomes smooth meanwhile keeping natural. 

On the other hand, we consider that because the ground truth is not directly but through the forward-backward consistency to penalize the forward optical flow, this makes the backward flow ($o_{eb}$ and  $o_{sb}$) serve as an effective intermidate to further ensure the smoothness of generated motions.

\subsection{Training Losses.}
We firstly train the Stage-I model, then using the trained Stage-I model to train Stage-II. The Stage-I model uses the same loss function as pix2pixHD~, except introducing a pixel loss between the sketches of synthesized images and the training sketches. Besides the $ \mathcal{L}_{sm} $, we also propose several other losses to train the stage-II model as follows.

A pixel-level loss is used between the input $v$ and output video $\hat{v}$ to reconstruct the training video, formulated as:
\begin{equation}
  \mathcal{L}_{pix} =   \mathcal{L}_1(\hat{v}-v)
\end{equation}

Following [As-rigid-aspossible stereo under second order smoothness priors., an unbiased second-order prior for high-accuracy motion estimation], a smoothness constraint loss $  \mathcal{L}_{osc}$ encourages the flow to be similar in a local neighborhood as follows,
\begin{equation}
  \mathcal{L}_{osc} =   \mathcal{L}_1( \bigtriangledown o_{xf} ) +    \mathcal{L}_1(\bigtriangledown o_{xb} ) ~~~~~~~, x\in{s,e}
\end{equation}

As mentioned before, we seperately apply a bidirectional frame reconstruction loss $  \mathcal{L}_{bfr}$ to the start \& end frame. For the start frame, the bidirectional frame reconstruction loss $ \mathcal{L}_{sbfr}$ can be formulated as
\begin{equation}
\begin{aligned}
      \mathcal{L}_{sbfr} = \sum_{t=1}^{N}{ m_{sf} \odot | v[0]- W(v[t],o_{sf}) |_1} \\
                            + \sum_{t=1}^{N} { m_{sb}\odot |v[t]-W(v[0],o_{sb}) |_1 }
\end{aligned}
\end{equation}
The end frame bidirectional reconstruction loss can be formulated as $\mathcal{L}_{ebfr}$:
\begin{equation}
\begin{aligned}
      \mathcal{L}_{ebfr} = \sum_{t=1}^{N} {m_{ef}\odot| v[N]- W(v[N-t],o_{ef}) |_1} \\
                            + \sum_{t=1}^{N} {m_{eb}\odot|v[N-t]-W(v[N],o_{eb}) |_1 }
\end{aligned}
\end{equation}

The forward-backward consistency loss $ \mathcal{L}_{fbc}$ for unmasked regions can be formulated as:
\begin{equation}
\begin{aligned}
  \mathcal{L}_{fbc} = \sum_{t=1}^{N} & m_{xf}[t]\odot |o_{xf}[t] - W(o_{xb}[t],o_{xf}[t]) |_1 \\
                &+ m_{xb}[t]\odot|o_{xb}[t] -W(o_{xf}[t],o_{xb}[t]) |_1 , ~~~~~~~x\in{s,e}
\end{aligned}
\end{equation}

\section{Experiments}
\label{sec:exp}
\subsection{Datasets}
As there is no standard sketch-to-video benchmark for use, we thus jointly utilize the public CelebAMask-HQ~\cite{Lee2020} dataset and VoxCeleb2~\cite{Chung2018} dataset to conduct experiments and evaluate our proposed method on the sketch-to-video task. We train our Stage-1 model on the CelebAMask-HQ~\cite{Lee2020} dataset to perform realisitic generation and train our Stage-2 model on the VoxCeleb2~\cite{Chung2018} dataset to better create temporal information for video synthesis. CelebAMask-HQ is a large-scale face image dataset with 30,000 high-resolution face images. We randomly select $30\%$ for testing and the rest for training. VoxCeleb2 is a famous dataset consisting of talking head videos (224p videos at 25fps) which has 1 million utterances for 6,112 celebrities. We use 145,569 videos for train and 4,911 for testing. Note that our validation samples in qualitative experiments are drawn by human artists.

\subsection{Implementation Details}
We get the sketch image by firstly performing edge detection via HED~\cite{Xie2017}. Then, the images in the CelebAMask-HQ dataset are further simplified by ~\cite{Simo-Serra2018} to mimic human drawing. Note that we only need the sketches of the first and last frame in videos and abandon others as input. The images and videos are rescaled to $256\times256$ pixels in all experimental settings.


We ask the participates to choose the video which one makes them feel more natural and gradual. We ask 10 particaptes to review each pair. 

\subsection{Qualitative Results}
\label{qual_res}
In the qualitative experiments, we propose several baselines, and conduct carefully designed experiments. We compare the performance and find out the real contribution of each stage of our model in this subsection. (Please zoom in to see the differences.)
\begin{figure*}[t]
  \includegraphics[width=\linewidth]{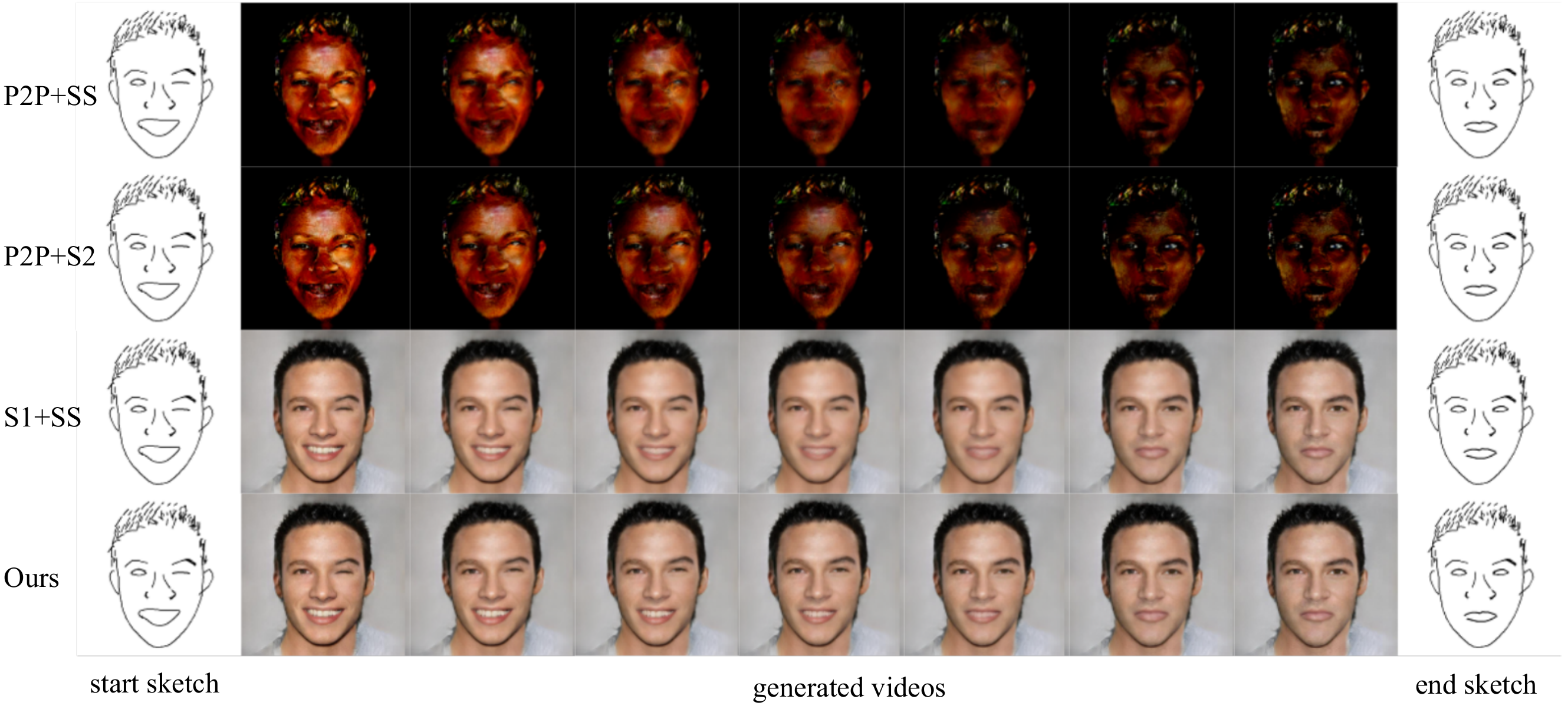}
  \caption{Qualitative Results. The S1 and S2 is corresponding to Stage-I and Stage-II module in our method. The P2P~\cite{Wang2018} and SS~\cite{Jiang2018} are both representative networks in image translation and video interpolation area. With the input start and end sketch, these four methods generate the video in the middle of this figure.}
  \Description{}
  \label{fig:qual_res}
\end{figure*}

\textbf{Baselines}
As this is the first work to create videos from given two rough hand-drawn sketches, we propose several baselines to compare current works with ours. 
\begin{itemize}
    \item \textbf{P2P+SS}~\cite{Wang2018}~\cite{Jiang2018}. Video interpolation is a classical vision problem aiming to synthesize high FPS videos from a low one, some of them can synthesize serval more frames from given two. The SuperSlomo~\cite{Jiang2018} (SS) is a famous video interpolation method which has been widely used in video creation. Meanwhile, the Pix2PixHD~\cite{Wang2018} (P2P) is one of the most famous image translation networks. We combine these two works to build a pipeline, in which the P2P method translates the input sketches into images, and the SS method caculates a linear flow between the two frames as input.
    \item \textbf{P2P+S2}~\cite{Wang2018}. To fairly compare the P2P+SS baseline with ours and in case the combination in P2P+SS makes the other slow down, we replace our Stage-I method with Pix2PixHD and keep our Stage-II (S2) method unchanged, which makes the comparison between S2 and SS a controlled experiment.
    \item \textbf{S1+SS}~\cite{Jiang2018}. For a more rigorous comparison of controlled experiments and compare the P2P with our the Stage-I model. In this setting, we combine our Stage-I method and SuperSlomo to find out the ability of video creation. Note the we modified the 

\end{itemize}

\textbf{Analysis of Qualitative Results.}
As shown in Fig.\ref{fig:qual_res}, all baselines with the P2P~\cite{Wang2018} as stage-I method produce a seriously distorted face in the generated videos. As we analyzed in Section.\ref{FRP}, due to the out-of-domain problem of rough hand-drawn sketches, the image translation model P2P~\cite{Wang2018} takes a sketch that has a free-form and different distribution not covered in the training datasets. This out-of-domain problem makes the result of the P2P seriously distorted. However, we find out all baselines with our Stage-I(S1) models have excellent and consistent semantics. We believe our Stage-I method alleviates the out-of-domain problem to a certain extent.

As to the terms of Stage-II, we mainly pay attention to the video generation performance. The generated qualities are found to depend on the choice of the Stage-I method. When stage-I is P2P, both the P2P+SS and the P2P+S2 are not good. The start and end frames are distorted and have a different identity. The results of both methods are gradually changing. We can see the P2P+SS baseline has a linear-like change, and the middle frames are blurry. However, one with our S2 module gradually changes the identity from the start to the end frame, and we can still see the human face organs of each frame, especially the eyes, which are lost in the P2P+SS baseline. In the S1+SS and our model(S1+S2), the difference is the teeth are buried in S1+SS(Please Zoom in to check the details.), where the motion is not linear to the underlying assumption of SS. Based on the above observation, we believe our stage-II method (S2) has better motion generation performance than the SS, and it could synthesize natural and gradual motion and has a learned prior.

\subsection{Quantitative Results}
\label{quan_res}

\label{abla_res}
\begin{figure*}[t]
  \includegraphics[width=\linewidth]{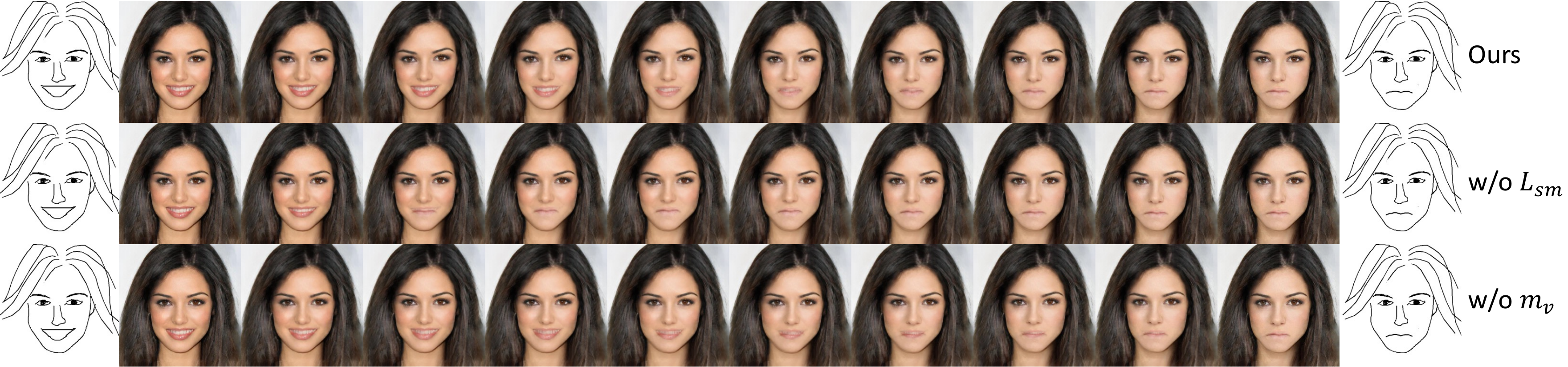}
  \caption{The ablation study of our model. The sketches from two ends are the input start and last sketches, repectively. The video frames inbetween are the generated video frames. The first row shows the results of our whole model, while the second raw removes the smooth motion loss of our model. In the third row, we remove the visible mask in the OFM Module.(Please zoom it to see the details.)}
  \Description{}
  \label{fig:abl}
\end{figure*}
\begin{figure*}
  \includegraphics[width=\linewidth]{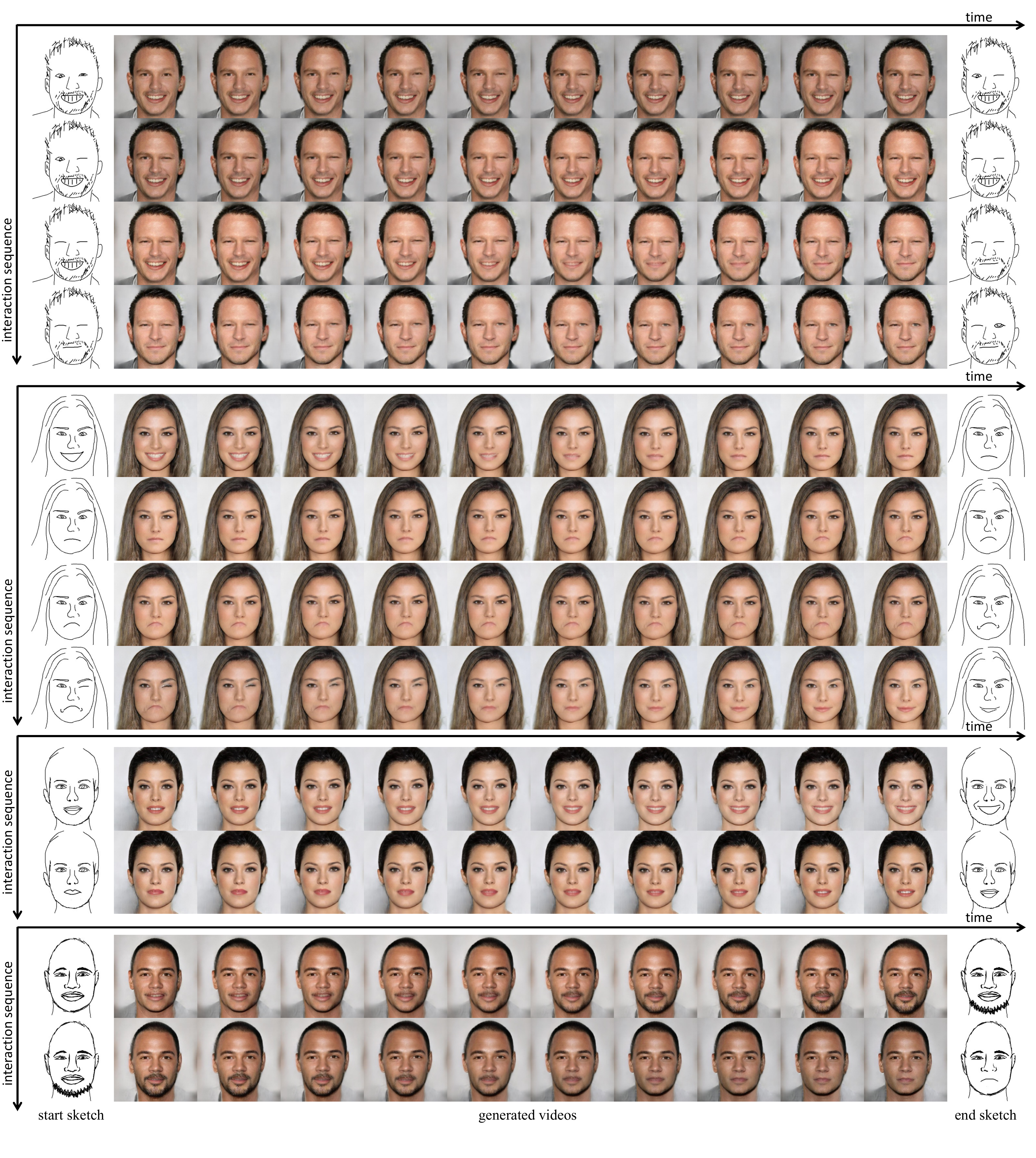}
  \caption{The Interaction results of our method. Each coordinate system has a one-time interaction process. The horizontal axis and vertical axis correspond to the interaction orders and the growth of frame index in one interact process, respectively. The first and last columns are the input start and end sketches, while the middle is the generated videos.}
  \Description{}
  \label{fig:interact}
\end{figure*}

\subsubsection{Evaluation Metrics.}
 For quantitative comparison, we evaluate the generation quality of output video. We adopt the \textbf{Fr\'echet Inception Distance}~\cite{Szegedy2016} metric. The FID measures the similarity between the two sets of images, and it is a widely used metric to evaluate the quality of generative results. When calculating the FID, we use sketches from edge detection and set the Stage-I module as P2P to evaluate the video quality and avoid the result affected by the out-of-domain problem.

Besides, we also employ two human subjective scores (HSS) to evaluate our methods. We divide the experiment into two sub-experiments. In the first part, we want to evaluate the quality of video generation, so we generate 100 samples from calculated sketches from the CelebA to evaluate the video quality. In the second part, we try to evaluate the ability of bad-drawn sketch based video generation. So, we generate another 100 samples from modified bad-drawn sketches for the same participants to check. We donote the first part as HSS\_w, and the second part as HSS\_r, according to whether their inputs are well-drawn sketches or rough ones. 

\begin{table}
\begin{center}
\begin{tabular}{l c c c c}
\toprule
Method &          HSS\_r $\uparrow$ & HSS\_w $\uparrow$ \\
\hline
 Ours / P2P~\cite{Wang2018}+SS~\cite{Jiang2018} & $1.0/0.0$ & $0.68/0.32$   \\
Ours / P2P~\cite{Wang2018}+S2 & $1.0/0.0$ & $0.47/0.53$\\
Ours / S1+SS~\cite{Jiang2018} & $0.64/0.36$ &  $0.77/0.33$  \\
\bottomrule
\end{tabular}
\end{center}
\caption{ Quantitative comparison of the pipelines in terms of the human subjective scores. }
\label{tab:hss}
\vspace{-1em}
\end{table}


\begin{table}
\begin{center}
\begin{tabular}{l c c c c}
\toprule
Method &      SS~\cite{Jiang2018} &  Ours  \\
\hline
 FID$\downarrow$  & $16.41$ & $15.63$ \\
\bottomrule
\end{tabular}
\end{center}
\caption{ Quantitative comparison of our module in terms of the FID score. }
\label{tab:hss}
\vspace{-1em}
\end{table}

\subsection{Ablation Study.}

We iteratively remove the smooth motion loss and visible mask to find out the actual function of them and whether these components work as expected or not. As shown in Fig.\ref{fig:abl}, the generated video of ours gradually and naturally changes from a big smile to pursed lips expression. However, without the smooth motion loss, the motion would be random and the big smile firstly suddenly becomes a snickering expression from the third to the fifth frame, then becomes a pursed lips expression. As we mentioned in Section.\ref{sec: sml}, this motion may be reasonable in the training dataset, but it makes the motion beyond the control of the given two sketches. As that hardly appears in the whole model, we believe that indicates the smooth motion loss works as expected.

\subsection{Interact Results.}
\label{interact_res}

To simulate the daily application scenario of video creation, we evaluate the interactive video synthesis ability of the model by conducting experiments as follows.
Firstly the user uses a sketching input as an initial sketch and then interactively changes various parts of the sketch, e.g., emotion, haircut, and facial organs. Our model synthesizes the video between every two sketches. 

The first coordinate system shows a casual way of face editing, and it shows the robustness of our method that can handle the regular video generation by changing expressions. The second coordinate system test some extreme expression changes, e.g., from the big smile face to the very sad face. The last two coordinate systems test entertaining video generation processes like growing or fading a beard or pout face. These interaction test results prove that our model has the potential ability to create video for normal users in practical use.

\section{Conclusion}
In this work, we propose a new sketch-to-video generation task from two arbitrarily drawn sketch images. Each sketch image is first divided into various parts carrying different attributes, goes through a feature retrieve and projection module to generate rich semantic features and the start or end frame. Unlike existing video generation works which rely on a good temporal guide video as auxiliary input, we abandon such guide input and propose a temporal motion prejection module to learn a normal distriobution for motions from various training videos. Such a distribution can be used to generate a motion variable in the testing phase, which is further fused with extracted semantic features via a developed feature blending module for feature decoding. The decoded flow images coupled with the start and end frame are finally used to synthesize the target video. Experiments on a combined facial dataset well validate the effectiveness of the proposed sketch-to-video approach. 


\bibliographystyle{ACM-Reference-Format}


\appendix









\end{document}